\documentclass[11pt]{article}

\usepackage[round]{natbib}
\usepackage{setspace}
\usepackage{geometry}
\usepackage[section]{placeins}
\usepackage[hidelinks]{hyperref}
\usepackage{graphicx}
\usepackage{xcolor}
\usepackage{titlesec}
\usepackage[page]{appendix}
\usepackage{enumerate}

\usepackage{subfiles}

\usepackage{lscape}
\usepackage{booktabs}
\usepackage{rotating}
\usepackage{multirow}
\usepackage{longtable}
\usepackage{caption}
\usepackage{subcaption}
\usepackage{float}
\usepackage{tabularx}
\usepackage{ragged2e}
\newcolumntype{Y}{>{\RaggedRight\arraybackslash}X}
\usepackage{pdflscape}
\usepackage{afterpage}

\usepackage{amsmath}
\usepackage{amssymb}
\usepackage{amsthm}
\usepackage{mathtools}
\usepackage{dsfont}

\usepackage[ruled,vlined]{algorithm2e}

\SetCommentSty{algcommstyle}

\usepackage{tikz}
\usetikzlibrary{bayesnet}

\usepackage{textcomp}
\usepackage{sourcecodepro}
\usepackage{listings}
\definecolor{commentgrey}{gray}{0.45}
\definecolor{backgray}{gray}{0.96}
\DeclareMathOperator*{\argmin}{arg\,min}

\DeclareMathOperator*{\argmax}{arg\,max}

\newcommand*{\cond}{\;\ifnum\currentgrouptype=16 \middle\fi|\;}

\newcommand*{\ttilde}{{\raise.17ex\hbox{$\scriptstyle\sim$}}}

\makeatletter
\newsavebox{\mybox}\newsavebox{\mysim}
\newcommand*{\distas}[1]{%
  \savebox{\mybox}{\hbox{\kern3pt$\scriptstyle#1$\kern3pt}}%
  \savebox{\mysim}{\hbox{$\sim$}}%
  \mathbin{\overset{#1}{\kern\z@\resizebox{\wd\mybox}{\ht\mysim}{$\sim$}}}%
}
\makeatother

\makeatletter
\def\moverlay{\mathpalette\mov@rlay}
\def\mov@rlay#1#2{\leavevmode\vtop{%
   \baselineskip\z@skip \lineskiplimit-\maxdimen
   \ialign{\hfil$\m@th#1##$\hfil\cr#2\crcr}}}
\newcommand*{\charfusion}[3][\mathord]{
  #1{\ifx#1\mathop\vphantom{#2}\fi\mathpalette\mov@rlay{#2\cr#3}}
  \ifx#1\mathop\expandafter\displaylimits\fi}
\makeatother

\newtheorem{theorem*}{Theorem}

\newtheorem{corollary*}[theorem*]{Corollary}

\newtheorem{proposition*}[theorem*]{Proposition}

\newtheorem{lemma*}[theorem*]{Lemma}

\theoremstyle{definition}

\newtheorem{definition*}{Definition}

\newtheorem*{remark}{Remark}

\newtheoremstyle{algodesc}{}{}{}{}{\bfseries}{.}{ }{}%
\theoremstyle{algodesc}

\newcommand\cD{\mathcal{D}}

\newcount\comments  
\newcommand{\genComment}[2]{\ifnum\comments=1{\textcolor{#1}{\footnotesize #2}}\fi}

\begin{document}

\title{Federated Learning via Synthetic Data}
\author{
    Jack Goetz \and
    Ambuj Tewari
}
\date{
    University of Michigan \\[2ex]%
    \today
}

\maketitle

\abstract{

Federated learning allows for the training of a model using data on multiple clients without the clients transmitting that raw data. However the standard method is to transmit model parameters (or updates), which for modern neural networks can be on the scale of millions of parameters, inflicting significant computational costs on the clients. We propose a method for federated learning where instead of transmitting a gradient update back to the server, we instead transmit a small amount of synthetic `data'. We describe the procedure and show some experimental results suggesting this procedure has potential, providing more than an order of magnitude reduction in communication costs with minimal model degradation.

}

\section{Introduction}

Federated Learning (FL) helps protect user privacy by transmitting model updates instead of private user data. However these updates could potentially be much larger than the private data they are replacing, and depending on the number of users each user may need to transmit updates multiple times during the training of a single model. This puts an increased communication cost on the user, and reducing that burden is an important research direction in federated learning \citep{kairouz2019advances, li2020federated, liu2020double}. We propose a training process which reduces the upload communication costs incurred by the user. This method was motivated by \citet{wang2018dataset}, which showed that training on large datasets can be fairly well approximated by specifically built small synthetic datasets (in that training on the small synthetic datasets can produce networks which are almost as good as ones trained on large datasets, as long as that training data is available when producing the synthetic data). We will build on this method to present a procedure which can \textbf{reduce the upload communication costs by one or two orders of magnitude}, while still producing good server models.

We will start by combining these ideas with ideas from data poisoning attacks to introduce the procedure at a high level. We will then discuss a few technical changes which make this different from either of those techniques, and which improve the performance of the procedure, including an extension of the procedure to reduce download communication costs as well as upload costs. We conclude with experiments and discuss some possible next steps in developing the procedure.

\section{Connection to Data Poisoning and Beyond}

\subsection{Motivation from data poisoning}

The inspiration for this method came from \citet{wang2018dataset}, but at a high level this method is also very similar to data poisoning attacks. In data poisoning an adversary wants to generate synthetic data such that when a model is trained using it, the model `does poorly' in some way. Let $f$ be our model (usually a neural network) with parameters $w$, where the model evaluated with parameters are denoted by $f(\cdot ;w)$. In \citet{munoz2017towards} they formulate this as a bi-level optimization problem:
\begin{flalign*}
    D_{po} \in \argmax_{D_{po}} &\quad \mathcal{L}(f(\cD_{te}(X); w), \cD_{te}(Y)) \\
    \text{s.t.} &\quad w \in \argmin_{w} L(f(\cD_{tr}(X) \cup D_{po}(X); w), \cD_{tr}(Y) \cup D_{po}(Y))\\
\end{flalign*}
where $D_{po}$ is the synthetic `poisoning' data, $\cD_{tr}$ is other training data and $\cD_{te}$ is test data, and the loss functions $\mathcal{L}, L$ are specified by the application (in standard data poisoning attacks they are usually some variant of the loss the model is fitting). The model trains on the training and poisoning data, gets model parameters $w$, which it believes are good, but which are actually bad.

If synthetic data can be generated to hurt the training process, then it could also be made to help it by flipping the argmax to an argmin. And we could use no other training data other than our synthetic data, giving us:
\begin{flalign*}
    D \in \argmin_{D} &\quad \mathcal{L}(f(\cD(X); w), \cD(Y)) \\
    \text{s.t.} &\quad w \in \argmin_{w} L(f(D(X); w), D(Y))\\
\end{flalign*}
where $\cD$ is our non-synthetic data and $D$ is our synthetic data. In its most simple form $L$ could be the quadratic approximation of the second order expansion of $\mathcal{L}$ at the current parameter location $w_0$, meaning that the optimization problem on the second level simply becomes a step of gradient descent (where we will allow the learning rate to be optimized over as well).

\begin{flalign*}
    D,\eta \in \argmin_{D, \eta} &\quad \mathcal{L}(f(\cD(X); w), \cD(Y)) \\
    \text{s.t.} &\quad w = w_0 - g\\
    &\quad g = \eta \nabla_{w} L(f(D(X); w_0), D(Y))\\
\end{flalign*}

In our federated optimization setting, during each training round the clients in the current cohort download the current model parameters $w_0$. The client then wants to find synthetic data $D, \eta$ such that $w$ has small loss on the client training data $\cD$. Once the client has generated the synthetic data, the client uploads this synthetic data to the server, which then gathers all client data together and does a single step of gradient descent. In many cases the model parameters $w$ may be much much larger in size than a single data point (or small number of data points), giving us a reduction (potentially quite significant) in \textbf{upload} transmission costs from client back to the server if we can upload a few synthetic data points instead of a gradient vector. Note this method is the same as in \citet{wang2018dataset}, just applied in the federated setting. We provide additions to the method to improve the effectiveness of the procedure and specializing it for FL. 

\subsection{Improvements}

We discuss several changes, all of which let to empirical improvements in the performance of our system. Some have intuition, but ultimately we were led by experiments. 

\subsubsection{Approximating standard federated learning}

We can change the upper optimization problem so that instead of operating directly on the training data, the synthetic data tries to approximate the gradient update the standard federated learning would have transmitted back to the server. We do this by first running the usual local update procedure for FL. This produces a `true update' $\theta$, which in standard FL we would transmit to the server directly. Now we adjust the upper optimization problem so that instead of fitting to the training data on the client, the synthetic data tries to induce an update $g$ which is similar $\theta$. In the above our $\mathcal{L}$ is now more generally some function of $\cD, w_0$ and $g$, so is $\mathcal{L}(\cD, w_0, g)$.

We tried two ways of inducing a similar update. The first was directly penalizing the difference between $\theta$ and $g$ with a simple squared loss $\mathcal{L} = ||\theta - g||^2$, where $\theta$ comes from updating our parameters from $w_0$ using our real data $\cD$. Here our loss function is working directly in parameter space. The other loss function $\mathcal{L}$ we tested was one based on the following procedure: Using the true update $\theta$, calculate the probability vectors $y'$ predicted by the updated network on each real data point. So $y' = f(\cD(X); w_0 - \theta)$. Then the overall loss function looks at the KL-div between the probability vector predicted by the true updated network, and the probability vector predicted by the induced updated network. This conceptually is trying to fit the induced update by its similarity in function space of the true updated network and the induced updated network. This worked well, but in the end the squared error on parameter space ended up being better. 

\remark
This may be due to the overparameterization of the network. There may be many different updates can create networks which are the same on a set of test points but differ across the function space. Even when the true data was augmented by randomly generated fake data (which corresponds to taking random samples of the functions) this was not as good as the squared loss (despite the appeal of fitting in function space). 

\subsubsection{Multiple steps of gradient descent}

There is no reason we need to limit ourselves to a single step of gradient descent. We can instead produce several batches of synthetic data $D_m$ for $m \in \{1,...,M\}$ which produce intermediate updates $g_m$. 

\begin{flalign*}
    \{D_m,\eta_m\}_{m=1}^M \in \argmin_{\{D_m,\eta_m\}_{m=1}^M} &\quad \mathcal{L}(\cD, w_0, g) \\
    \text{s.t.} &\quad w = w_0 - g\\
    &\quad g = \sum g_m\\
    &\quad g_m = \eta_m \nabla L(f(D_m(X); w_{m-1}), D_m(Y))\\
\end{flalign*}

This was already done in \citep{wang2018dataset}, where they found this to be useful. We found that it was not just helpful, but absolutely vital in approximating standard FL updates which had taken multiple steps of SGD. Although at first this seems like it would be computationally expensive as it would require computation of Hessians, a well known trick from \citet{pearlmutter1994fast} allows us to do this efficiently.

\subsubsection{Normalizing intermediate and final updates}

We also moved to using normalized SGD (so we calculate the gradient, normalize it and then multiply by the learning rate). Conceptually this mean the synthetic data $D_m$ only defines the direction of the update $g_m$, and the learning rate $\eta_m$ completely defines the magnitude of the update. This enhanced the stability of our procedure. 

We also normalized again after all $M$ intermediate steps to produce our induced update $g$, and the `learning rate' $H$ which we multiply this by is equal to the norm of the true update $\theta$. The normalization of the overall update $g$ ensures that we produce updates which are the same norm as the true update. This is valuable from an optimization perspective (conceptually this acts as a side constraint, feeding additional information to guide the optimization procedure). 

\subsubsection{Trainable $Y$ in synthetic data}

Our experiments focused on the classification setting. In the original paper \citep{wang2018dataset} synthetic data was given a fixed class, so only the covariates were synthetic. This was done since having a label which is fractional did not really make any sense. However neural networks do not naturally work on categorical data space, they naturally work in the probability distribution space (the neural network will produce a probability of each class for a given data point and then we reinterpret that into a class label). However we noticed that fixing these labels was very limiting, so we allowed our synthetic data to have synthetic label distributions. So each synthetic data point has covariates $X$ which are in the same space as our real data, and a label vector which are probabilities of each class instead of a single class label. Similar methods were shown in \cite{sucholutsky2019soft} to be equally powerful. We project and normalize during training to keep all probabilities between 0 and 1 and summing to 1. We still use the standard cross entropy loss (or KL loss since they are the same here), but now our label vectors are no longer one hot vectors, instead are generally dense. This type of adjustment cannot generally be made in data poisoning attacks since the attacker needs to create data in the same space as the model. However here the client and server are working together, and so the server can adjust it's own training process to accommodate this super-labelled synthetic data. From here on we will consider the output of our neural network $f$ to be the probability vector, as opposed to the class which is the argmax of that vector.

\section{Synthetic data generating procedure}

We present the subroutines for each server model update step. The most important part is the $updateFromSynthetic$ function, which dictates how the synthetic data is decoded into an update. There are two functions $localUpdate$ and $aggregate$ which are undefined. These are the `standard' methods from FL, so $localUpdate$ is likely several passes of SGD over the client data, and $aggregate$ could be federated averaging or something more advanced. 

\begin{algorithm}[ht!]
\caption{Server update} \label{alg:server_update}
  \SetAlgoLined
  \KwIn{Clients in cohort $\mathcal{C}$, current server model parameters $w_0$}
  \KwOut{Server model update $g$}
  
  \For{$c \in [1:|\mathcal{C}|]$}{
    Transmit current model parameters $w_0$ to client $\mathcal{C}[c]$.\;
    Client runs $clientUpdate$ and transmits back synthetic data $(D_c, H_c)$
    $g_c = updateFromSynthetic(D_c, w_0, H_c)$.\;
  }
  $g = aggregate(\{g_c\})$.\;
  return $g$.\;
  Remark: Here we have been a little sloppy with notation. The subscripts on $g_c$ enumerate over the clients in the cohort, but the values transmitted back by each client are the final $g$ in Algorithm \ref{alg:updateFromSynthetic}.
\end{algorithm}

\begin{algorithm}[ht!]
\caption{clientUpdate} \label{alg:clientUpdate}
  \SetAlgoLined
  \KwIn{Transmitted model parameters $w_0$, client data $\mathcal{D}$, distillation learning rate $\alpha$}
  \KwOut{Synthetic data $D = \{D_m, \eta_m\}_{m=1}^M$, norm of the true local update $H$}
  $\theta = localUpdate(\mathcal{D}, w_0)$\;
  $H = ||\theta||$\;
  Initialize $D^0$ \;
  \For{$t \in [1:T]$}{
  \textit{Forward}\;
    $g = updateFromSynthetic(D^{t-1}, w_0, H)$\;
    Evaluate loss $\mathcal{L} = \sum (\theta[i] - g[i])^2$\;
    \textit{Backwards}.\;
    \For{$m \in [1:M]$}{
    $D^{t}_m = D^{t-1}_m - \alpha \nabla_{D_m} \mathcal{L}$\;
    $\eta^{t}_m = \eta^{t-1}_m - \alpha \nabla_{\eta_m} \mathcal{L}$.\;
    }
  }
  $D = D^T$.\;
  return $D, H$
\end{algorithm}

\begin{algorithm}[ht!]
\caption{updateFromSynthetic} \label{alg:updateFromSynthetic}
  \SetAlgoLined
  \KwIn{Synthetic data $D$, initial model parameters $w_0$, true update norm $H$}
  \KwOut{Induced update $g$}
  \For{$m \in [1:M]$}{
    $\tilde{g_m} = \nabla_{w} L(f(D_m(X); w_{m-1}), D_m(Y))$\;
    $g_m = \eta_m \frac{\Tilde{g_m}}{||\tilde{g_m}||}$\;
  }
  $\tilde{g} = \sum\limits_{m=1}^M g_m$\;
  $g = H \frac{\tilde{g}}{||\tilde{g}||}$\;
  return $g$\;
\end{algorithm}

\newpage

\begin{figure}[ht!]
	\centering
	\begin{tabular}{cc}
		\includegraphics[width=0.99\textwidth,height=0.25\textheight,clip = true]{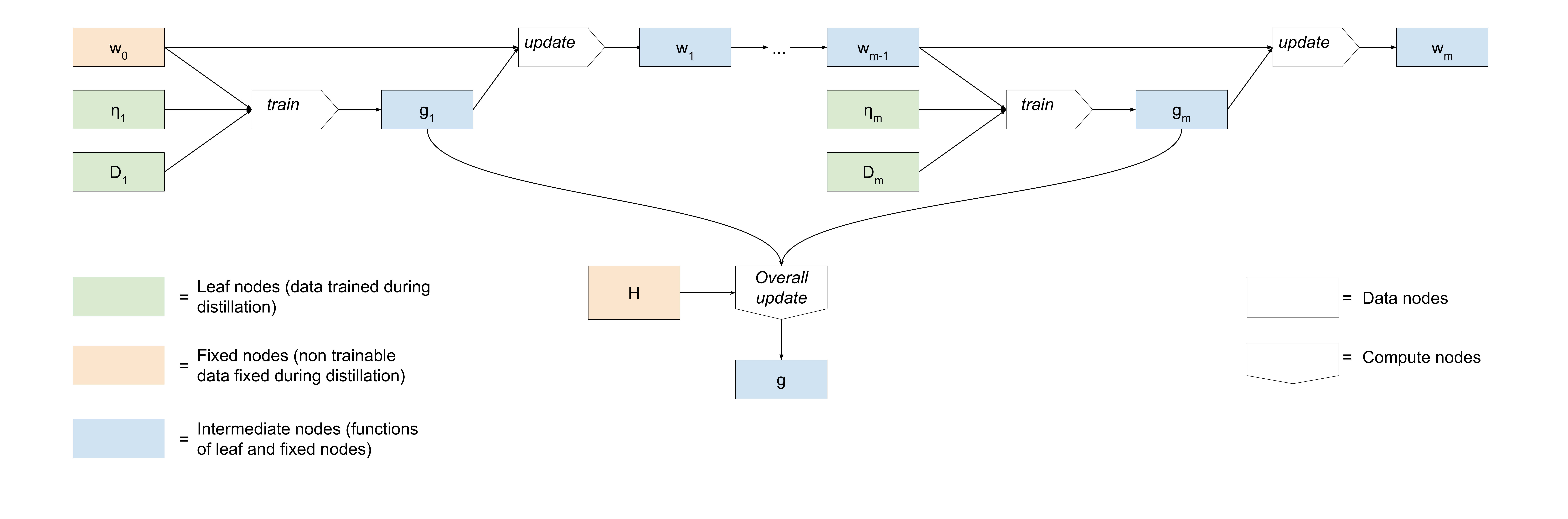}\\
	\end{tabular}
	\caption{Computational graph for $updateFromSynthetic$. Compute node formulas: \textit{train}: $g_m = \eta_m \frac{\Tilde{g_m}}{||\tilde{g_m}||}$ where $\tilde{g_m} = \nabla L(f(D_m(x); w_{m-1}), D_m(Y))$ where $L(y, z) = D_{KL}(z||y)$. \textit{update}: $w_{m+1} = w_m - g_m$. \textit{Overall update}: $g = H \frac{\tilde{g}}{||\tilde{g}||}$ where $\tilde{g} = \sum\limits_{m=1}^M g_m, H = ||\theta||$.}
	\label{fig:comp_graph}
\end{figure}



In Figure \ref{fig:comp_graph} the green leaf nodes comprising of $(D_1, \eta_1...D_m, \eta_m)$ and the node $H$ are what will be send back to the server, and the server runs $updateFromSynthetic$ to produce $g$, which it will then treat as if that were the gradients uploaded by the client directly. It is vital that the $updateFromSynthetic$ on the client and on the server are exactly the same! Figure \ref{fig:comp_graph} provides the computational graph used to derive the backprop computations required. Although it may seem computationally prohibitive since we need the Hessian of our parameters, they are only needed in a vector-Jacobian product, and fortunately this can be efficiently calculated in $\mathcal{O}(n)$ using a well known trick from \citet{pearlmutter1994fast}. 

An important structure we use here is that it is possible for $D_i = D_j$. The advantage of this is we only have to transmit that synthetic data once, but we generate multiple $g_i$ from it. Conceptually it is similar to `training on the synthetic data for multiple epoch'. We have found this to be very powerful. The changes to the computational graph above would just be to change the current $D_m$ leaf nodes in the graph into intermediate nodes, and have a set of leaf data nodes which can point to multiple of the current leaf nodes.

\subsection{Tracking the best approximation}

Ideally we would produce synthetic data such that $g = \theta$, but in practice this will almost never happen. And our loss function $\mathcal{L}$ is directly on the parameter space, which is not ideal since it does not take into account the impact small changes in parameters have on the output of the neural network. To try and take this into account, we adapt a technique from non-convex optimization. In non-convex optimization one common heuristic is to, separately from the sequence of solutions produced during your optimization procedure, keep track of the best solution, and only update that if you get a better solution (even if you let the optimization procedure move to a worse solution in the hope that it will eventually find an even better one). For example when training a NN you can use a hold out set, and keep track of the parameters which perform best on the hold out set, even if you let the parameters be updated to worse (on the hold out set) parameters. Usually people use the same objective function which they are optimizing (or same except evaluated on a different dataset). However here we will use two very different objective functions. 

After every round, to test the quality of our induced update $g$, we calculate the cross entropy loss on our the client's training data set. We keep track of the synthetic data which induces the lowest cross entropy loss, using this loss to define our `best' synthetic data. This is a pretty standard technique, but here we are using this to implicitly `fit' to two complimentary (but not identical) objective functions: we want an induced update which is close to the true update, and which performs well on our training data. This is of course completely ad hoc, and almost certainly could be improved. However empirically this helped stabilize our optimization procedure.

\section{Experiments}

We test our synthetic data approach with experiments emulating federated learning. Our experiment setup is based on the MNIST experiments in \cite{mcmahan2016communication}, using the same MNIST CNN network structure (further experiments with CIFAR are in the works).  We used the same data sharding scheme for the iid and non-iid data (100 clients, random sharding for the iid data, 2 classes per client with 300 points each for non-iid data). For the $localUpdate$ we use SGD with 5 epochs, batch size of 10 and learning rate of 0.02, and for $aggregate$ on the server we use federated averaging, with a cohort size of 10. 

For the distillation parameters (parameters that affect the synthetic data) we used 5 batches of synthetic data, each containing 10 synthetic data points, and $updateFromSynthetic$ will train on those batches for 5 epochs (giving us an $M=25$), with a distillation learning rate $\alpha$ of 0.2. We also used Adam to train the synthetic data in Algorithm \ref{alg:clientUpdate} (as opposed to the GD stated). 

This procedure requires transmission of 50 synthetic data points and the norm of $\theta$, requiring transmission of 39701 floats, or just 2.4\%, or $\frac{1}{40}^{th}$ of the 1663370 floats required to transmit a gradient update for the MNIST CNN model. We train the synthetic data for 300 updates, requiring about 5.5 times the computation required for just running the $localUpdate$ on it's own. Note that because we always run the same $localUpdate$ as a first step, our synthetic data FL will always be more computationally expensive.

The selection of these distillation parameters involved a small amount of tuning (constrained by what we considered to be reasonable computational and communication costs), but could almost certainly be tuned more to improve the results. At the coarse level we tuned, both the iid and non-iid partitioning schemes used the same tuning parameters (where as one would expect them to have different optimal parameters at a more granular level of tuning). 

\subsection{Quality of procedure on IID and non-IID clients}

\begin{figure}[ht!]
	\caption{Comparing synthetic data FL to full gradient transmission FL}
	\centering
	\begin{tabular}{cc}
	    iid client partitioning\\
		\includegraphics[width=0.99\textwidth,height=0.35\textheight,clip = true, trim= {80 50 20 0} ]{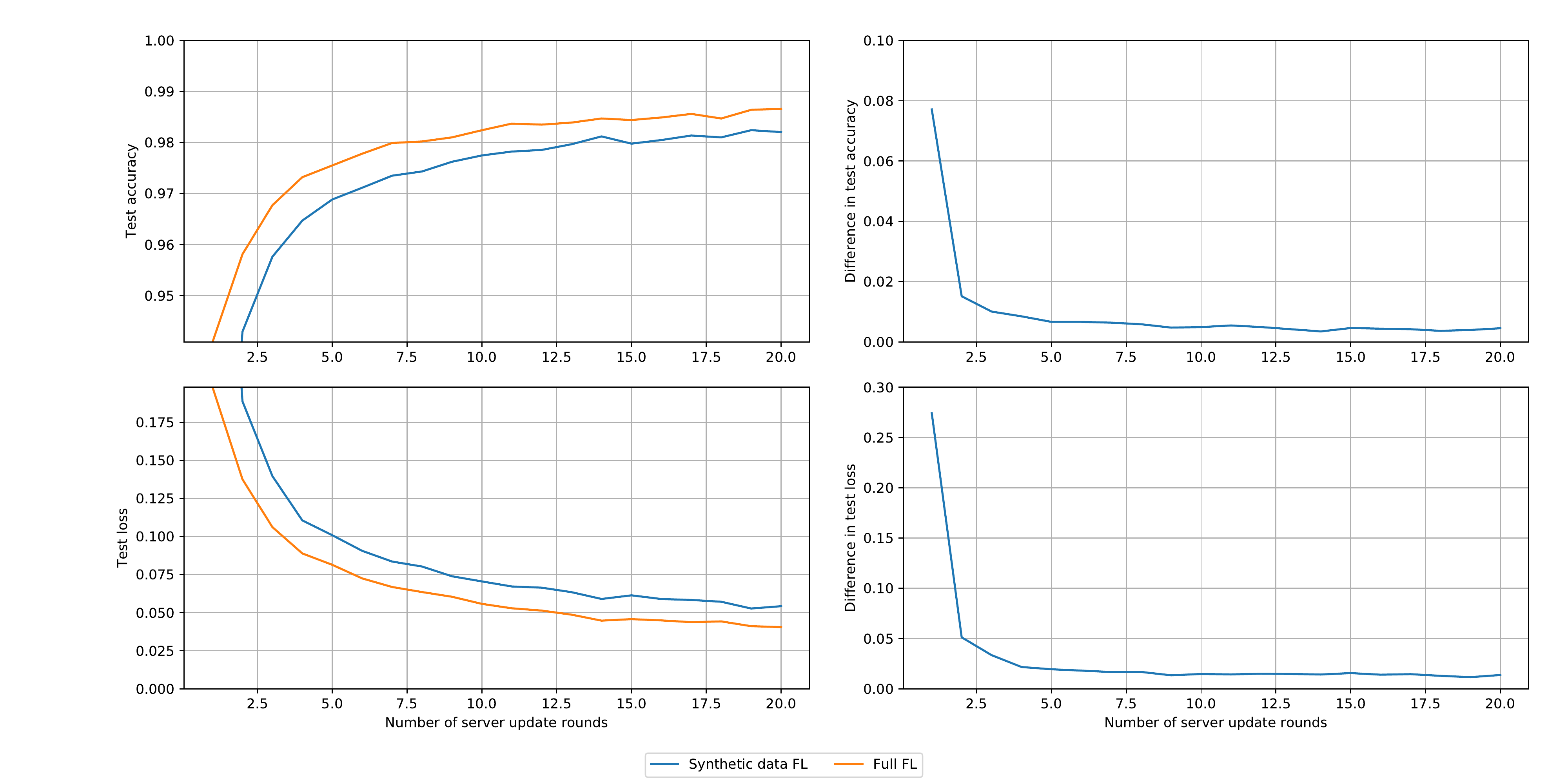}\\
		Non-iid client partitioning\\
		\includegraphics[width=0.99\textwidth,height=0.35\textheight,clip = true, trim= {80 0 20 0} ]{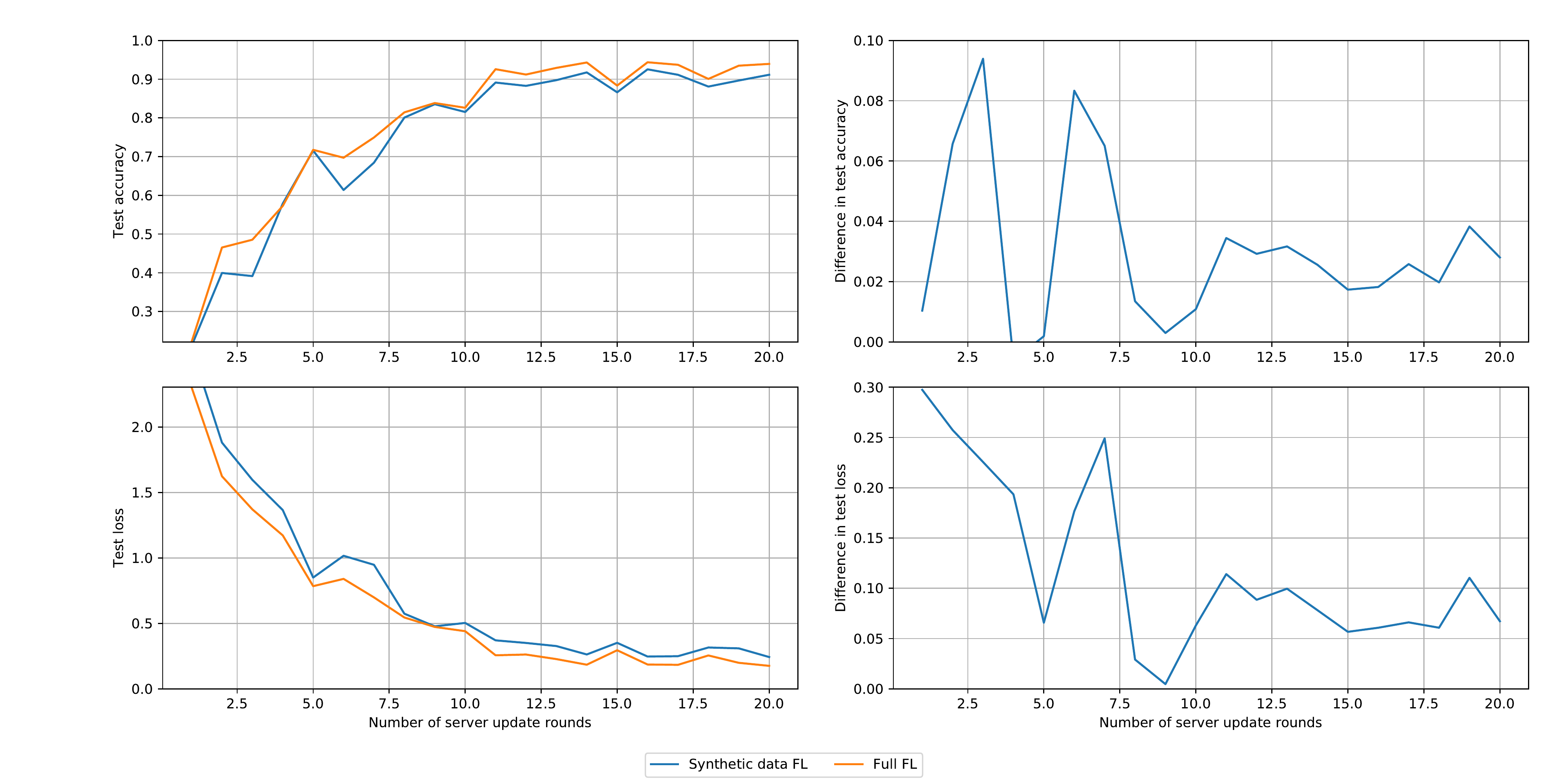}\\
	\end{tabular}
	\label{fig:compare}
\end{figure}

In these experiments the seeds which dictate how the data is partitioned between users and randomness with the FL procedure (which client is in which cohort etc) are all the same, though they are different from the seeds used to select tuning parameters. This means the standard FL procedure is the same for all runs. And the seed used for creating the synthetic data (initialization etc) are different. The plots on the left show raw values, and the plots on the right show the difference between using synthetic data FL and standard full gradient transmission FL. The first two rows in Figure \ref{fig:compare} show results for the iid data partitioning, while the second two show the non-iid partitioning.

The synthetic data FL appears comparable to the full gradient transmission FL, while only requiring a small fraction of the upload transmission costs. 

\subsection{Robustness to distillation learning rate}

\begin{figure}[ht!]
	\caption{Testing different learning rates}
	\centering
	\begin{tabular}{cc}
		\includegraphics[width=0.99\textwidth,height=0.35\textheight,clip = true, trim= {80 0 20 0} ]{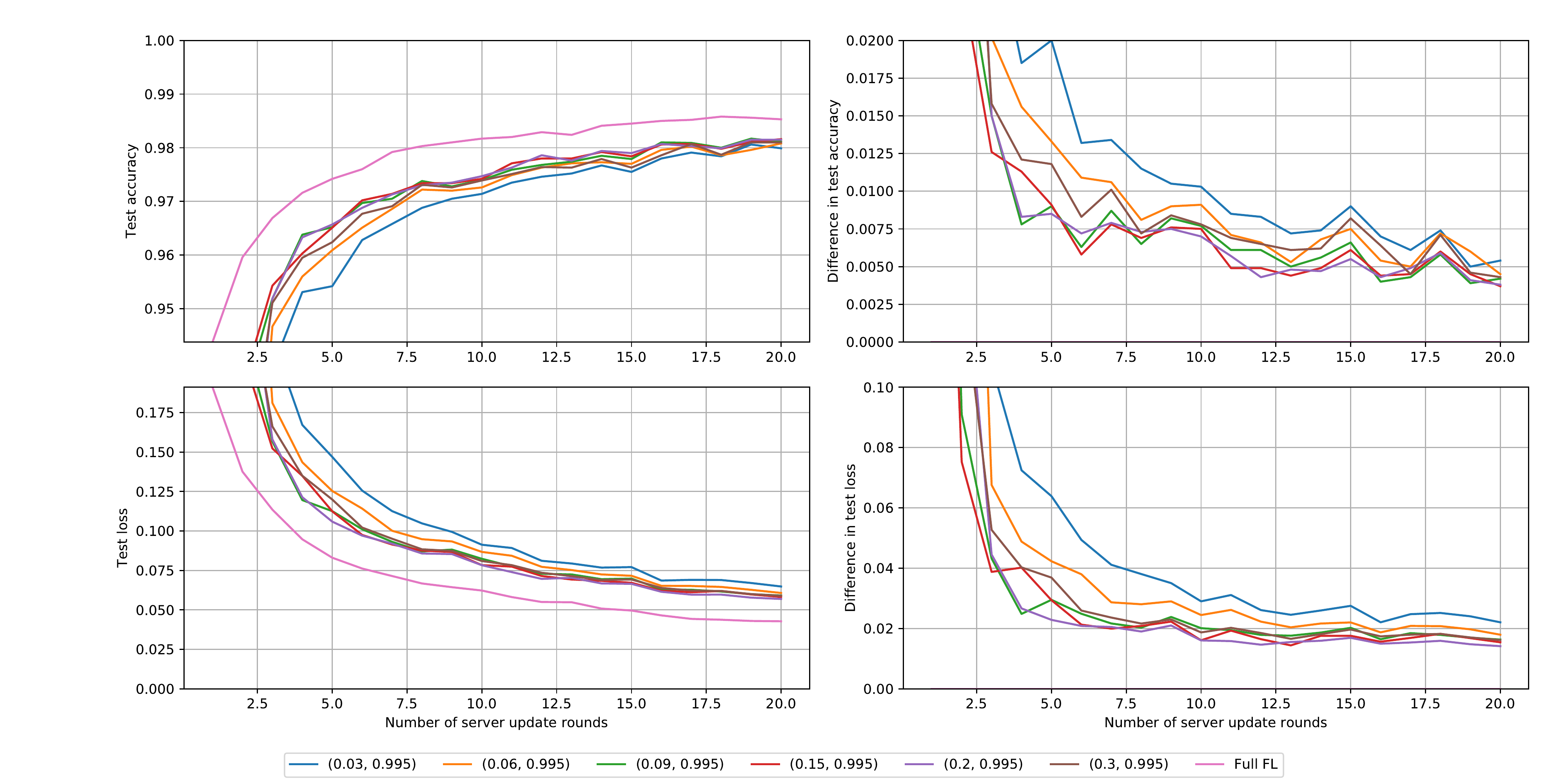}\\
	\end{tabular}
	\label{fig:lr_robustness}
\end{figure}

The biggest cost independent tuning parameter (in that changing this value does not change the computational or communication costs) is the distillation learning rate $\alpha$. Figure \ref{fig:lr_robustness} suggests that we are not super sensitive to the value of this learning rate, as most rates between 0.03 and 0.3 produced very similar values (the 0.995 is the learning rate decay, and had even less of an effect than the learning rate). Since hyperparameter tuning is challenging in FL this robustness is very valuable. 

\subsection{Trade off between communication, compute and approximation quality}

Unsurprisingly as you increase the number of synthetic data points you use, or the amount of client compute you use to generate that synthetic data, you change your approximation quality and correspondingly how similar to full FL you perform. However there are several parameters you can change to increase the computation or the communication. For example if you are willing to double your communication costs, you can either double the size of each synthetic batch, or double the number of synthetic batches (or some mixture of the two). Note that both of these also increase your compute, so you may need to account for that. Similarly if you are willing to double your computation (without changing communication) you can double the amount of time you spend training your synthetic data, or you can double the number of epochs over the synthetic data you use in $updateFromSynthetic$. It appears that all of these suffer from diminishing returns, and so it is best to use some amount of all of these methods. There may also be a connection to the number of steps of SGD taken during $localUpdate$ (one might expect it to be better to have more synthetic batches if there are more steps of SGD, see below), but we do not yet fully understand this.

\subsection{Adapt for server to client transmission}

The above method is for client to server transmission, but does nothing for server to client. However we can also use this to transmit an approximation of the server model to the client. We need to know the initial state of the model on each client, and this can be achieved if we fix our model initialization protocol and then transmit the seed first used to initialize on the server to the client. Then we send synthetic data which induces an update that bring a model with that known initialization close to the current model on the server. The transmission and use of synthetic data is the same as our current setup (just sending synthetic data from server to client instead). 

Although the principle is the same, there are differences which make this direction more challenging. The biggest difficulty is now we are replicating thousands (and possibly much more) of updates with just a few updates, as opposed to in client to server, where we are using a few updates to replicate tens of updates. In particular as the training process evolves, we need to replicate a growing number of updates, as opposed to the client to server direction where the number of updates we need to approximate is (roughly) constant. We believe this to be at least one of the main reasons why the approximation for the reverse direction is much more challenging computationally. 

Experimentally we found using synthetic data to transmit the model in both directions to be much more challenging. The biggest issue we encountered were the occurrence of complete failures, where the server failed to approximate the model at all and the synthetic data ends up inducing a model which is no better than the initialized model. We were able to overcome this issue via a brute force approach, where we initialized the synthetic data using multiple seeds, separately trained each one and used the data which induced the model which was most similar to the current server model. Although this multiplicatively increases the server computational costs, in our setup the server is owned by the researchers and so server compute is much less valuable than client compute or communication costs. And since this training can be done in parallel, this can be done without dramatically increasing the wall time of our training procedure. 

Initial experimental results suggest that this server to client transmission is viable, although would benefit from further refinement. We repeated used the same experimental setup as above, with the exception that we used synthetic data to transmit the required models in both directions. For the server to client transmission, we used 100 synthetic data points in 10 batches of 10 data points (requiring only $\frac{1}{20}^{th}$ the download cost compared to the full model parameters), trained for 600 updates, and tried 10 seeds for each server initialization. 

\begin{figure}[ht!]
	\caption{Synthetic data transmission in both directions}
	\centering
	\begin{tabular}{cc}
		\includegraphics[width=0.99\textwidth,height=0.35\textheight,clip = true, trim= {80 0 20 0} ]{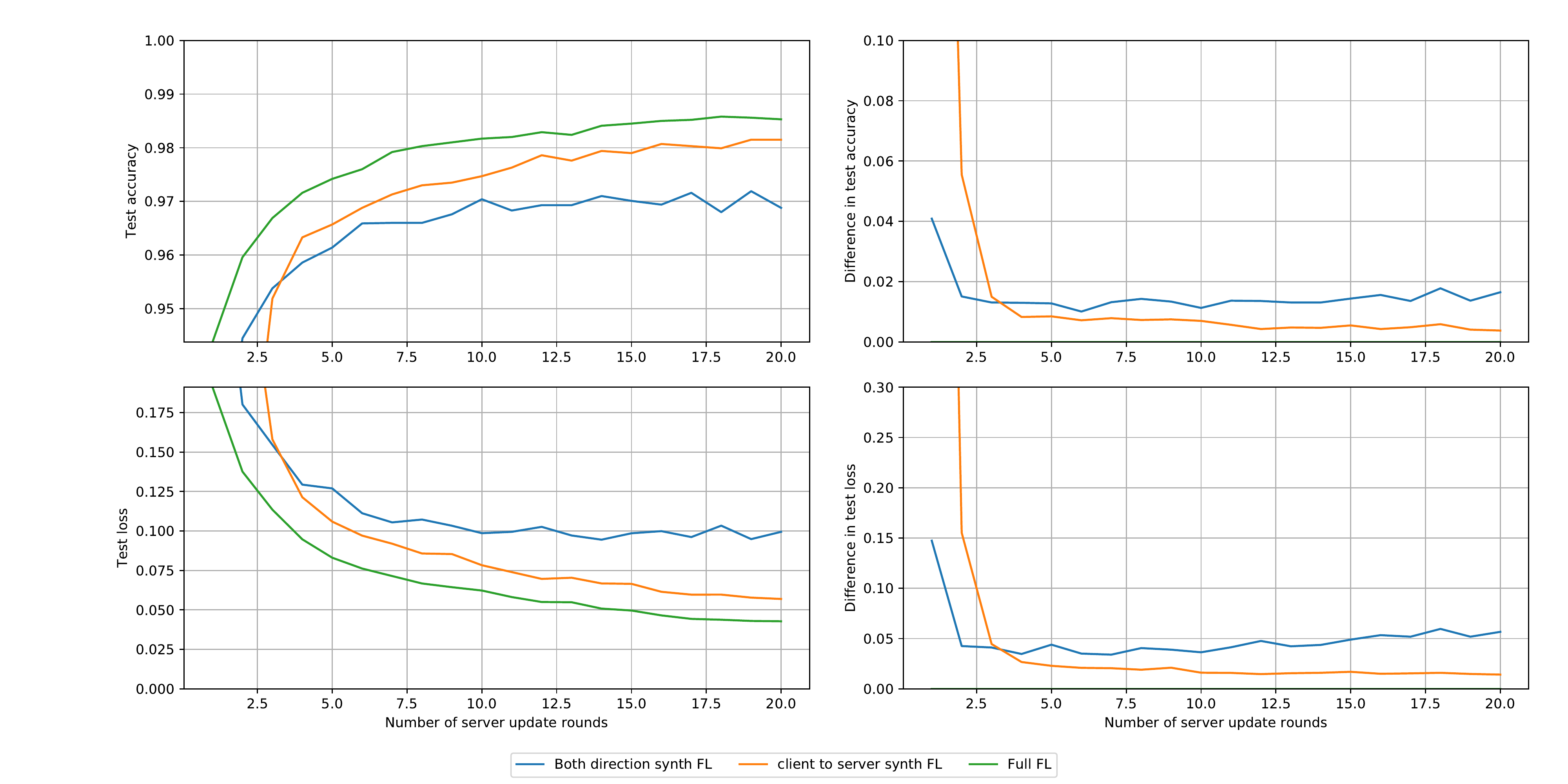}\\
	\end{tabular}
	\label{fig:double_distill}
\end{figure}

The double synthetic data FL is able to train a good model, trading 1.5\% loss in accuracy for over 90\% reduction in download communication costs. However it appears that the model may be unable to further improve. Overcoming this shortfall is critical and the subject of continuous work.

\section{Possible Future Work}

These initial results seem very promising, but there are still many things we need to understand before we know if this will actually be useful in practice. Below we have several of the directions we think are most promising. 

\subsection{Async updates}

This is probably the most speculative but also most exciting possible future direction. In \citep{wang2018dataset} they show that the current system is extremely sensitive to $w_0$ being the same on the client and on the server. However they also showed that if you train over a distribution of $w_0$'s on the client (so during each synthetic data update you draw from $p(w_0)$ and use that for the update), and then the server draws from that same distribution, you can get back good performance. This hints that we might be able to use a similar procedure to generate synthetic data which is robust to producing good updates even if the model parameters have changed a little. You probably need to train on $w_0$ drawn uniformly from some epsilon ball around the transmitted $w_0$, and who knows how much compute you will need, or whether you can do this effectively on a large enough ball (might need clipping of your server update, and even that might not be enough). So there is a lot of uncertainty here. But given that something similar was shown to work in \citep{wang2018dataset} it seems like it is worth trying. 

\subsection{Privacy concerns}

Transmitting gradients has the benefit that they are much harder to interpret than sending data (though of course they are not impossible to learn from). There is the risk that the synthetic data we transmit may be much more revealing about the data on the client. In \cite{wang2018dataset} on the MNIST data set they produced synthetic data which looked very `real', in that you could clearly recognize the synthetic data as showing the number of the label. Since our data no longer has single labels this might not be an issue, but we need to look into it possibly solve it. One possible solution would be to project our probabilities away from 0 to prevent any synthetic data point from representing a real label. We could also learn our synthetic data in a differentially private way, possibly adding noise before the approximation process.  

\subsection{Experiments with heterogeneous client resources}

In FL we often have that different clients have very different quality hardware available. Clearly we can adapt our distillation parameters to accommodate that, using fewer synthetic data points for clients with less bandwidth and more compute on clients with faster phones. The question is whether this is a good idea, especially if there is correlation between resource usage and the type of data on the client. Does increasing resource usage on clients who can afford it lead to a strictly better model (where the improvement may be focused on areas where the faster clients have more data)? Or does this lead to detrimental experiences for the slower clients by biasing our models? 

\subsection{Remove needing $localUpdate$}

We found that fitting to the true FL update worked better than creating synthetic data using the training data directly. However we may just have done it poorly. Removing the need for a $localUpdate$ could be valuable when we are working in very low compute environments. 

\subsection{Analysis from a compression view}

Here we are really just using the `training' procedure of the neural network as a decompressor, where the synthetic data is the compressed true local update $\theta$. Can arbitrary vectors be equally well compressed, or is there something about the connection between the decompressor and the nature of the vectors being compressed? Intuitively one might hypothesize that the manifold of possible updates is much smaller than all of $\mathbf{R}^d$ (where $d$ is the number of parameters in the model), and that they contain some sort of `nested' property as you reduce the amount of data used to train. That could explain why we can well approximate lots of training with lots of data, using little training with little data.

\section{Acknowledgements}

We would like to thank Mikhail Yurochkin for his helpful advice and insights in both guiding the project and presenting this work.

%
%
%
%
%
%
%
%
%
%

\bibliographystyle{apalike}
\bibliography{refs}

\end{document}